\documentclass{article}

\usepackage[preprint]{neurips_2024}

\usepackage{natbib}
\setcitestyle{numbers,square}
\usepackage{tikz}
\usepackage{graphicx}
\usepackage{wrapfig}
\usepackage{booktabs}
\usepackage{adjustbox}
\usepackage{multirow}
\usepackage{xspace}
\usepackage{array}
\usepackage{bm}
\usepackage{bbm}
\usepackage{amsthm,amsmath,amssymb}
\usepackage{algorithm}
\usepackage{algorithmic}

\usepackage{listings}
\usepackage{wrapfig}
\usepackage{subcaption}
\usepackage{caption}
\usepackage{url}
\usepackage{colortbl}
\usepackage{adjustbox}
\usepackage{makecell}
\usepackage{pifont}
\usepackage{setspace}
\usepackage{fancybox}
\usepackage{tocloft}
\usepackage{etoc}
\usepackage{hyperref}
\usepackage{colortbl}
\definecolor{darkblue}{rgb}{0, 0, 0.5}
\hypersetup{colorlinks=true, citecolor=darkblue, linkcolor=darkblue, urlcolor=darkblue}
\usepackage[most,skins,theorems]{tcolorbox}
\usepackage{cleveref}
\tcbset{
  aibox/.style={
    width=\linewidth,
    top=8pt,
    bottom=4pt,
    colback=blue!6!white,
    colframe=black,
    colbacktitle=black,
    enhanced,
    center,
    attach boxed title to top left={yshift=-0.1in,xshift=0.15in},
    boxed title style={boxrule=0pt,colframe=white,},
  }
}
\newcolumntype{C}[1]{>{\centering\let\newline\\\arraybackslash\hspace{0pt}}m{#1}}
\newtcolorbox{AIbox}[2][]{aibox,title=#2,#1}

\newcommand{\name}{GUI-R1 }

\title{GUI-R1: A Generalist R1-Style Vision-Language Action Model For GUI Agents}

\author{
Run Luo$^{1,2}$ \ \ Lu Wang$^{3}$ \ \ Wanwei He$^{1,2}$ \ \ Longze Chen$^{1,2}$ \ \ Jiaming Li$^{1,2}$ \ \ \\
Min Yang$^{1,2}$ \ \ Xiaobo Xia$^{3}$
\\
\small$^1$Shenzhen Institute of Advanced Technology, Chinese Academy of Sciences \\
\small$^2$University of Chinese Academy of Sciences \\
\small$^3$National University of Singapore\\
\sc\small \{r.luo@siat.ac.cn \ \ \ m.yang@siat.ac.cn \ \ \ xiaoboxia.uni@gmail.com\}
}

\begin{document}

\maketitle

\begin{abstract}
Existing efforts in building graphical user interface (GUI) agents largely rely on the training paradigm of supervised fine-tuning (SFT) on large vision-language models (LVLMs). However, this approach not only demands extensive amounts of training data but also struggles to effectively understand GUI screenshots and generalize to unseen interfaces. The issue significantly limits its application in real-world scenarios, especially for high-level tasks. Inspired by reinforcement fine-tuning (RFT) in large reasoning models (e.g., DeepSeek-R1), which efficiently enhances the problem-solving capabilities of large language models in real-world settings, we propose GUI-R1, the first reinforcement learning framework designed to enhance the GUI capabilities of LVLMs in high-level real-world task scenarios, through unified action space rule modeling. By leveraging a small amount of carefully curated high-quality data across multiple platforms (including Windows, Linux, MacOS, Android, and Web) and employing policy optimization algorithms such as group relative policy optimization (GRPO) to update the model, GUI-R1 achieves superior performance using only 0.02\% of the data (3K vs. 13M) compared to previous state-of-the-art methods like OS-Atlas across eight benchmarks spanning three different platforms (mobile, desktop, and web). These results demonstrate the immense potential of reinforcement learning based on unified action space rule modeling in improving the execution capabilities of LVLMs for real-world GUI agent tasks. 
The codebase is available at \url{https://github.com/ritzz-ai/GUI-R1.git}. 

\end{abstract}

\begin{figure}[h]
\centering
\begin{subfigure}[b]{0.32\textwidth}
    \includegraphics[width=\textwidth]{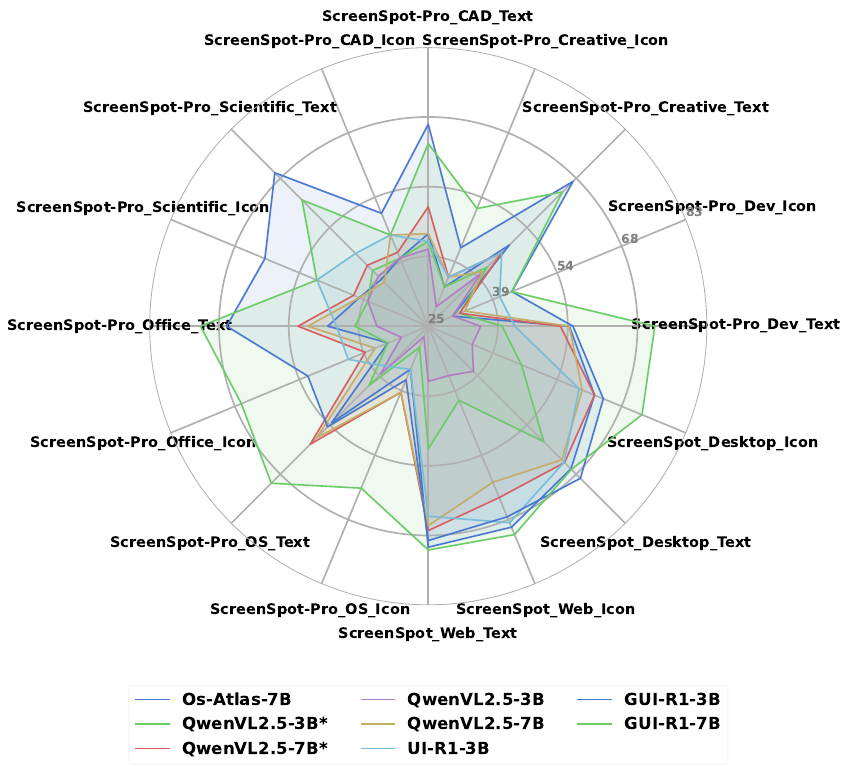}
    \caption{Grounding capability.}
    \label{fig:subfig_a}
\end{subfigure}
\hfill
\begin{subfigure}[b]{0.32\textwidth}
    \includegraphics[width=\textwidth]{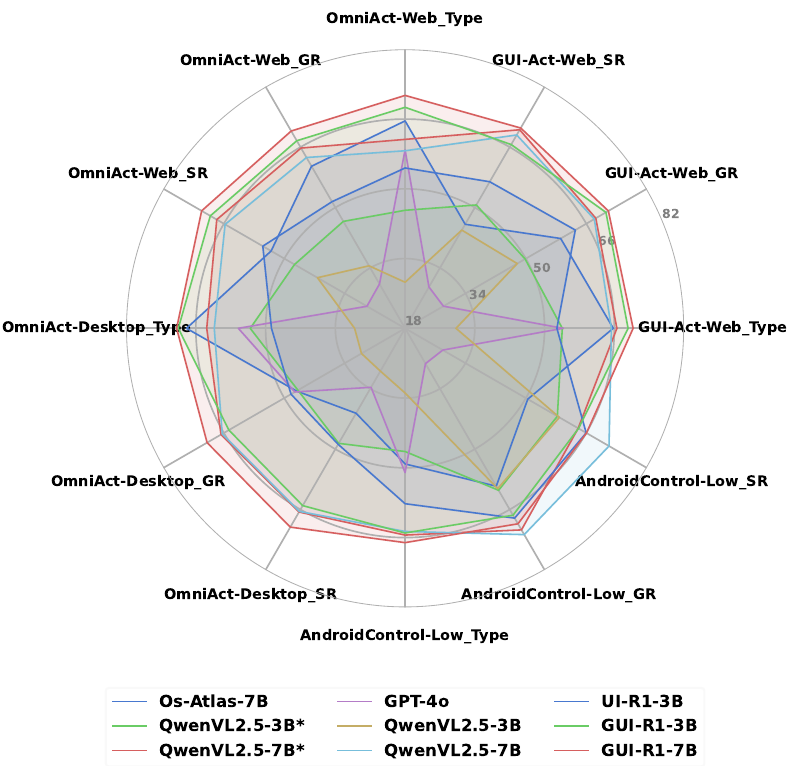}
    \caption{Low-level task capability.}
    \label{fig:subfig_b}
\end{subfigure}
\hfill
\begin{subfigure}[b]{0.32\textwidth}
    \includegraphics[width=\textwidth]{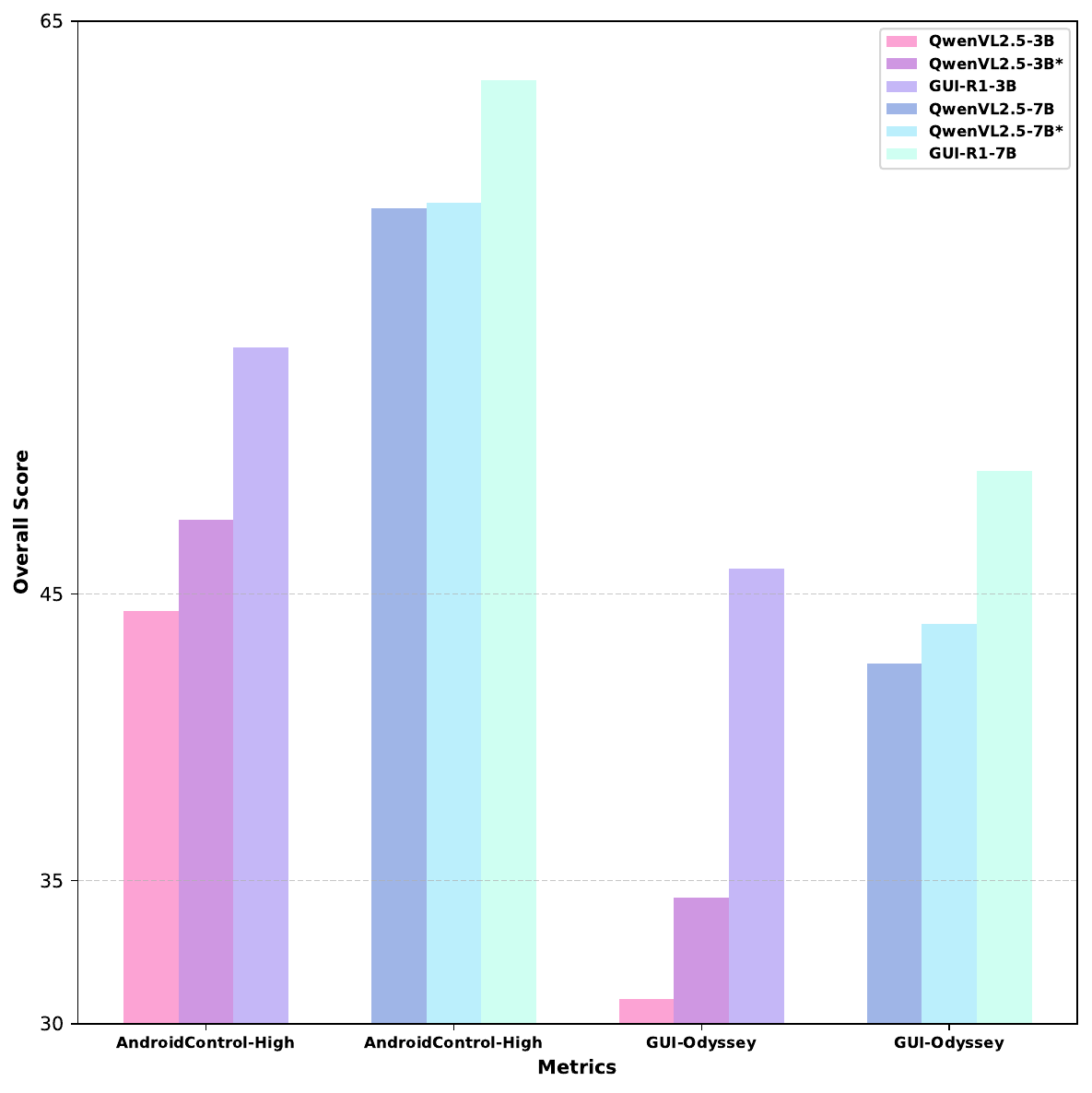}
    \caption{High-level task capability.}
    \label{fig:subfig_c}
\end{subfigure}
\caption{\name achieves the best performance on eight evaluation datasets covering various platforms and task granularities, demonstrating the promising potential of RFT in GUI agent tasks.}
\label{fig:main_performance}
\end{figure}

\section{Introduction}
Recent studies~\cite{osatlas,uitars,seeclick} have explored the use of large vision-language models (LVLMs)~\cite{qwen2.5vl} to develop graphical user interface (GUI) agents capable of performing high-level complex tasks. These agents analyze the screen as a self-contained source of information for decision-making, without relying on environment-based textual descriptions such as HTML or accessibility trees. This approach offers greater flexibility in agent decision-making. However, previous works have predominantly relied on the training paradigm of supervised fine-tuning (SFT), which not only requires large amounts of high-quality training data but also struggles to effectively comprehend GUI screenshots and generalize to unseen interfaces. These limitations have significantly hindered the real-world applicability of these works, particularly for high-level GUI tasks that lack explicit step-by-step instructions.

Rule-based reinforcement fine-tuning has recently emerged as an efficient and scalable alternative to SFT, requiring only a small number of examples to fine-tune models effectively while demonstrating strong performance and generalization capabilities in domain-specific tasks. RFT has been increasingly adopted for developing various LVLMs~\cite{visual-rft,visionr1,r1v,vlmr1,ui-r1}. Inspired by these advancements, this study extends the rule-based reinforcement learning~(RL) paradigm to the domain of GUI agents, which focuses on GUI action prediction tasks within a unified action space driven by high-level instructions. Specifically, LVLMs generate multiple responses (trajectories) for each input, containing both reasoning traces and final answers. These responses are evaluated using a unified action space reward function designed in this work, and the model is updated through policy optimization~\cite{deepseekr1}. This iterative self-learning process enhances the model's reasoning capabilities in action prediction and its generalization to out-of-distribution (OOD) scenarios.
By modeling a unified action space, we efficiently curate high-quality data spanning multiple platforms, including Windows, Linux, MacOS, Android, and Web, while avoiding action prediction conflicts across different platforms. 

As demonstrated in Figure~\ref{fig:main_performance}, the proposed framework~(GUI-R1) achieves superior performance using only 0.02\% of the data (3K vs. 13M) compared to previous state-of-the-art methods like OS-Atlas~\cite{osatlas} across eight benchmarks covering three different platforms (mobile, desktop, and web) and three levels of task granularity (low-level grounding, low-level tasks, and high-level tasks). Before delving into details, we clearly emphasize our
contribution as follows.
\begin{itemize}
    \item We propose GUI-R1, the first framework that utilizes rule-based reinforcement fine-tuning to enhance the reasoning capabilities of LVLMs in high-level GUI action prediction tasks.
    \item We design a rule-based unified action space reward function, which efficiently validates GUI task responses across different platforms and task granularities. This ensures reliable and efficient data selection and model training.
    \item Leveraging the rule-based unified action space reward function, we construct GUI-R1-3K, which is a high-quality fine-tuning dataset with diversity and complexity. This dataset significantly improves both training efficiency and model performance.
    \item We conduct a comprehensive evaluation of GUI agents, covering three distinct platforms (desktop, mobile, and web) and three levels of task granularity (low-level grounding, low-level tasks, and high-level tasks) across eight benchmarks. Experimental results demonstrate that our GUI-R1 is leading in multiple realistic cases. This creates a strong baseline of GUI agents for future research.    
    %As shown in Figure 1, \name achieves substantial performance improvements over the previous SOTA model, OS-Atlas, while using only 0.02\% of the data. This demonstrates its potential as an alternative to the SFT training paradigm for developing future GUI agents.
\end{itemize}

\section{Related Work}
\subsection{GUI Agents}
Autonomous agents driven by large foundation models (e.g., large language models (LLMs) and large vision-language models~(LVLMs)) have gained significant attention for their powerful interactive capabilities~\cite{sumers2023cognitive}. These operating systems via programs or API calls~\cite{wang2024survey,sun2023corex}. However, the closed-source nature of most commercial software limits access to internal APIs or code, which promotes a shift in research toward GUI agents. Different from traditional programmatic agents, GUI agents simulate human interactions via mouse and keyboard inputs, which enable broader flexibility in solving complex tasks. Recent works have advanced this direction. For instance, UGround~\cite{uground} developed a specialized GUI grounding model for precise GUI element localization.  OS-Atlas~\cite{osatlas} introduced large action models to handle general agent tasks by interpreting human intentions and predicting actions in the form of function calls. UITars~\cite{uitars} proposed a more comprehensive method by combining GUI-related pretraining with task-level reasoning fine-tuning to better capture the complexity of GUI interactions.  
Nevertheless, these methods all rely on the paradigm of supervised fine-tuning (SFT), which suffers from two main limitations: (1) the training process requires vast amounts of diverse data; (2) the models exhibit limited generalization capabilities, which struggle to understand GUI screenshots and adapt to unseen interfaces.  These limitations motivate the development of a more advanced learning paradigm for GUI agents beyond traditional SFT methods.

\subsection{Reinforcement Fine-Tuning}
Rule-based reinforcement fine-tuning,   exemplified by OpenAI o1~\cite{jaech2024openai} and DeepSeek-R1~\cite{deepseekr1}, has demonstrated strong performance in mathematical reasoning~\cite{shao2024deepseekmath}, code generation~\cite{code-r1}, and multi-step logic tasks~\cite{RAGEN}. Subsequent studies have extended this paradigm to multimodal models by designing task-specific reward functions for vision-based tasks, such as correct class prediction in image classificati~\cite{pan2025metaspatial,r1v,meng2025mm}, intersection-over-union (IoU) metrics in image localization and detection~\cite{visionr1,visual-rft}, and accurate click position prediction in low-level GUI grounding tasks~\cite{ui-r1}. These works demonstrate that verifiable reward signals, e.g., symbolic correctness or execution-based feedback, can effectively substitute traditional supervision.  
Despite the strong potential of RFT in various tasks, it remains underexplored in complex high-level GUI agent tasks. Compared to other domains, building intelligent agents for high-level GUI tasks is particularly challenging due to diverse UI layouts, implicit task semantics, and long-horizon action dependencies. This imposes higher demands on the model's contextual learning and understanding capabilities.  
To the best of our knowledge, \name is the first RFT-based framework specifically designed for high-level GUI agents.

\section{GUI-R1 Framework}

\begin{figure}[h]
    \centering
    \includegraphics[width=0.98\textwidth]{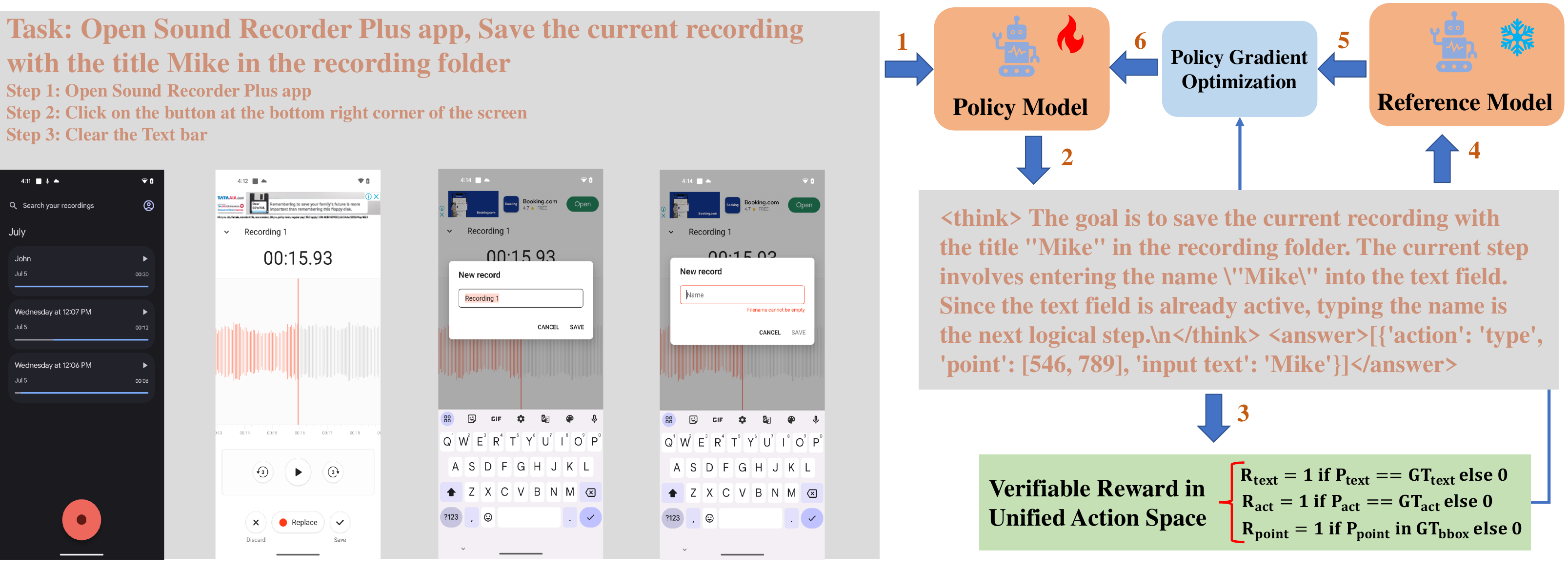}
    \caption{\textbf{Overview of the \name Framework.} Given the high-level instruction, action history, and visual image inputs, the policy model generates multiple responses containing reasoning steps. Then the verifiable rewards, such as action type reward, click point reward, and input text reward, are used with the policy gradient optimization algorithm to update the policy model.}
    \label{fig:framework}
\end{figure}

\name is based on a reinforcement learning training paradigm designed to enhance the ability of GUI agents to complete sophisticated instructional tasks. As shown in Figure~\ref{fig:framework}, unlike low-level tasks, high-level GUI tasks lack explicit and fine-grained instructions, which require action predictions based on high-level task objectives and execution history. This imposes greater demands on the model's contextual learning and understanding capabilities.
\subsection{Preliminaries}

We define the goal of GUI agents in high-level instructional tasks as understanding and executing low-level instructions to complete the high-level task $Q$, based on the current interface image $I$, and the execution history $H$. Formally, given the input $Q$, $I$, and $H$, the model generates a set of candidate responses $O = \{o_1, o_2, \dots, o_N\}$, where each response contains attributes of the predicted low-level action $o^{\text{act}}$, input text $o^{\text{text}}$, and input point $o^{\text{point}}$. Each response is evaluated using a unified action space reward function to compute its reward $\{r_1, r_2, \dots, r_N\}$. GRPO~\cite{deepseekr1} is applied to estimate advantages and update the policy model under KL divergence constraints. The relative advantage $A_i$ of the $i$-th response is calculated as follows:

$$
A_i = \frac{r_i - \texttt{mean}(\{r_1, r_2, \dots, r_N\})}{\texttt{std}(\{r_1, r_2, \dots, r_N\})},
$$

where $\texttt{mean}$ and $\texttt{std}$ denote the mean and standard deviation of the rewards, respectively.

\subsection{Verifiable  Rewards in Unified Action Space}

We adopt a unified action space modeling strategy, which extracts action space categories across different platforms and integrates them into a unified action space. This ensures that all high-level instructions can be decomposed into a sequence of atomic actions, resolving action space conflicts in multi-platform data joint training. Based on the unified action space, we design verifiable reward functions to evaluate the accuracy of predicted actions to guide reinforcement learning. We detail these verifiable rewards below.

\noindent \textbf{Format reward.} Following previous work~\cite{meng2025mm,deepseekr1,visionr1}, we introduce format rewards during training to evaluate whether the generated output adheres to the expected structural format, including both syntactic and semantic validity. Specifically, format rewards guide the model to generate reasoning processes and final answers in a structured format, which play a critical role in self-learning and iterative improvement during reinforcement fine-tuning. The format reward templates used in training and inference are as follows, where `$<$think$>$' represents the reasoning process and `$<$answer$>$' represents the final answer.
\begin{AIbox}{Unified Action Space Prompt for Task Training and Inference}
You are GUI-R1, a reasoning GUI Agent Assistant. In this UI screenshot $\boldsymbol{<image>}$, I want you to continue executing the command $\boldsymbol{task}$, with the action history being $\boldsymbol{history}$.

Please provide the action to perform (enumerate from [\texttt{complete}, \texttt{close/delete}, \texttt{press\_home}, \texttt{click}, \texttt{press\_back}, \texttt{type}, \texttt{select}, \texttt{scroll}, \texttt{enter}]), the point where the cursor is moved to (integer) if a click is performed, and any input text required to complete the action.

Output the thinking process in $<$$\boldsymbol{think}$$>$ $<$$/\boldsymbol{think}$$>$ tags, and the final answer in $<$$\boldsymbol{answer}$$>$ $<$$/\boldsymbol{answer}$$>$ tags as follows:
$<$$\boldsymbol{think}$$>$ ... $<$$/\boldsymbol{think}$$>$ $<$$\boldsymbol{answer}$$>$[{`action': enum[\texttt{complete}, \texttt{close/delete}, \texttt{press\_home}, \texttt{click}, \texttt{press\_back}, \texttt{type}, \texttt{select}, \texttt{scroll}, \texttt{enter}], `point': [x, y], `input\_text': `no input text [default]'}]$<$$/\boldsymbol{answer}$$>$.
\end{AIbox}
\begin{AIbox}{Unified Action Space Prompt for Grounding Training and Inference}
You are GUI-R1, a reasoning GUI Agent Assistant. In this UI screenshot $\boldsymbol{<image>}$, I want you to continue executing the command $\boldsymbol{task}$, with the action history being $\boldsymbol{history}$.

Please provide the action to perform (enumerate from [\texttt{click}]), the point where the cursor is moved to (integer) if a click is performed, and any input text required to complete the action.

Output the thinking process in $<$$\boldsymbol{think}$$>$ $<$$/\boldsymbol{think}$$>$ tags, and the final answer in $<$$\boldsymbol{answer}$$>$ $<$$/\boldsymbol{answer}$$>$ tags as follows:
$<$$\boldsymbol{think}$$>$ ... $<$$/\boldsymbol{think}$$>$ $<$$\boldsymbol{answer}$$>$[{`action': enum[\texttt{click}], `point': [x, y], `input\_text': `no input text [default]'}]$<$$/\boldsymbol{answer}$$>$.
\end{AIbox}

\noindent \textbf{Accuracy rewards.} For the model's predicted output $o = \{o^{\text{act}}, o^{\text{text}}, o^{\text{point}}\}$, which consists of three components: $o^{\text{act}}$ (action type, e.g., click, scroll), $o^{\text{point}}$ (click point position), and $o^{\text{text}}$ (input text), we define the accuracy reward $R_{\text{acc}}$ as a combination of action type reward $R_{\text{act}}$, click point reward $R_{\text{point}}$, and input text reward $R_{\text{text}}$, i.e., $R_{\text{acc}} = R_{\text{act}} + R_{\text{point}} + R_{\text{text}}.$ This design provides reliable correctness rewards for all actions.

\noindent \textbf{Action type reward.} The action type reward $R_{\text{act}}$ is calculated by comparing the predicted action type $o^{\text{act}}$ with the ground truth action type $gt^{\text{act}}$. If $o^{\text{act}} == gt^{\text{act}}$, the reward is 1; otherwise, it is 0. This simple yet effective evaluation mechanism guides action type prediction.

\noindent \textbf{Click point reward.} The click point reward $R_{\text{point}}$ is calculated by comparing the predicted click point $o^{\text{point}} = [x, y]$ with the ground truth bounding box $gt^{\text{bbox}} = [x_1, y_1, x_2, y_2]$. The calculation formula is as follows:

$$
R_{\text{point}} = 
\begin{cases} 
1 & \text{if } o^{\text{point}} \in gt^{\text{bbox}}, \\ 
0 & \text{otherwise}.
\end{cases}
$$

\noindent \textbf{Input text reward} The input text reward $R_{\text{text}}$ is calculated by comparing the predicted input text $o^{\text{text}}$ with the ground truth text parameter $gt^{\text{text}}$ using the semantic $F_1$ score. The calculation formula is as follows:

$$
R_{\text{text}} = 
\begin{cases} 
1 & \text{if } F_{1}(o^{\text{text}}, gt^{\text{text}}) > 0.5, \\ 
0 & \text{otherwise}.
\end{cases}
$$

\noindent \textbf{Response reward.} The final response reward is composed of format rewards and accuracy rewards, defined as:$R_o = \alpha R_{\text{f}} + \beta R_{\text{acc}},$ where $R_{\text{f}}$ represents the format reward, $R_{\text{acc}}$ represents the accuracy reward, and $\alpha$ and $\beta$ are weighting parameters respectively.

\subsection{Training Data Curation}
\noindent \textbf{Data collection.} We collect data related to GUI tasks from multiple open-source datasets, including FineWeb~\cite{fineweb}, UIBert~\cite{uibert}, AMEX~\cite{amex}, RICOSCA~\cite{li2020mapping}, as well as portions of pretraining data from Seeclick~\cite{seeclick} and OS-Otlas~\cite{osatlas}. This leads to $\sim$14M examples of grounding and low-level task data. Additionally, we collect $\sim$30K high-level GUI data points from OS-Otlas instruction datasets. In total, we gather $\sim$14M examples spanning multiple platforms (including Windows, Linux, MacOS, Android, and Web) and various task granularities (grounding, low-level, and high-level).

\begin{figure}[h]
\centering
\begin{subfigure}[b]{0.49\textwidth}
    \includegraphics[width=\textwidth]{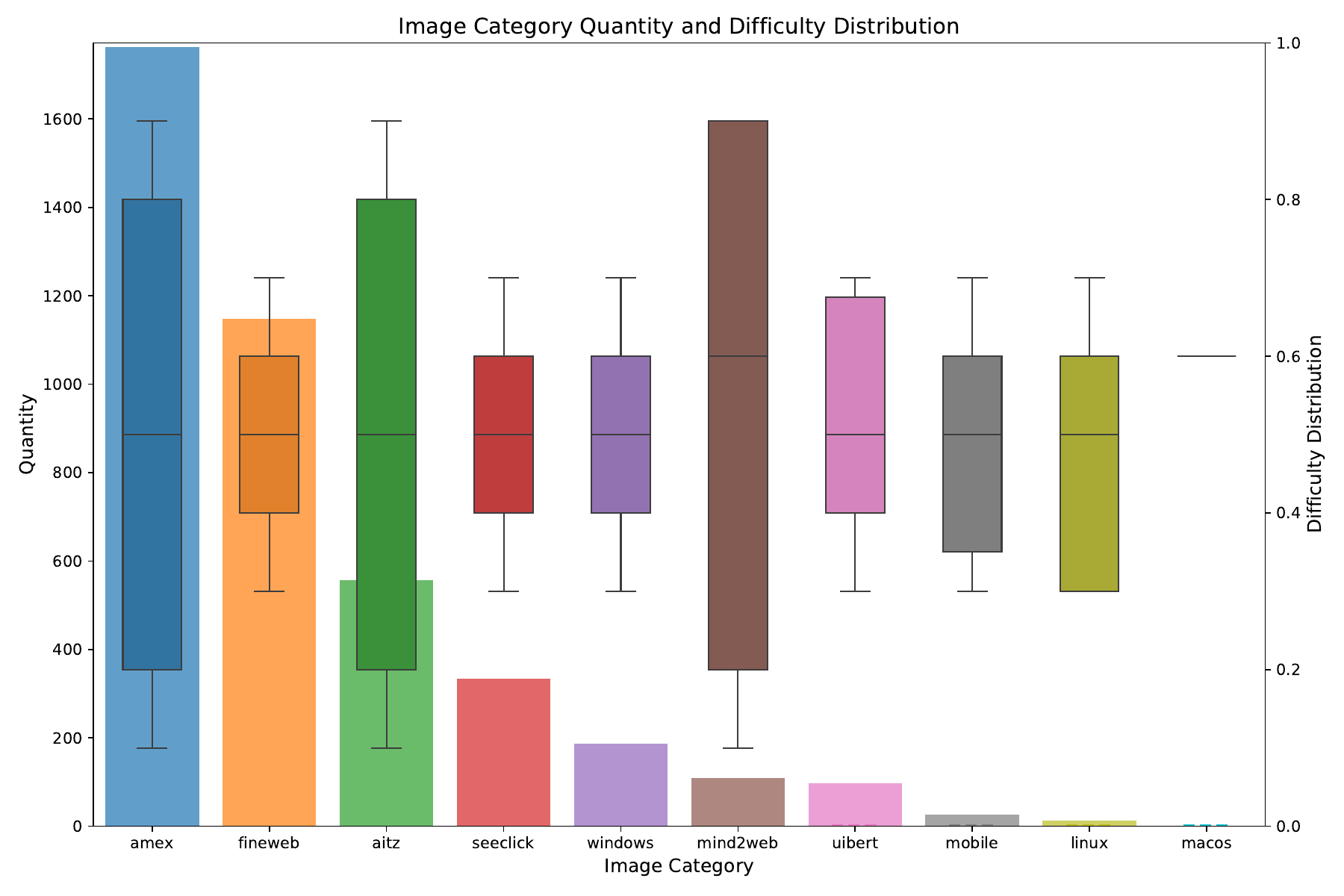}
    \caption{\scriptsize Image category quantity and difficulty distribution.}
    \label{fig:image_data_dist}
\end{subfigure}
\hfill
\begin{subfigure}[b]{0.49\textwidth}
    \includegraphics[width=\textwidth]{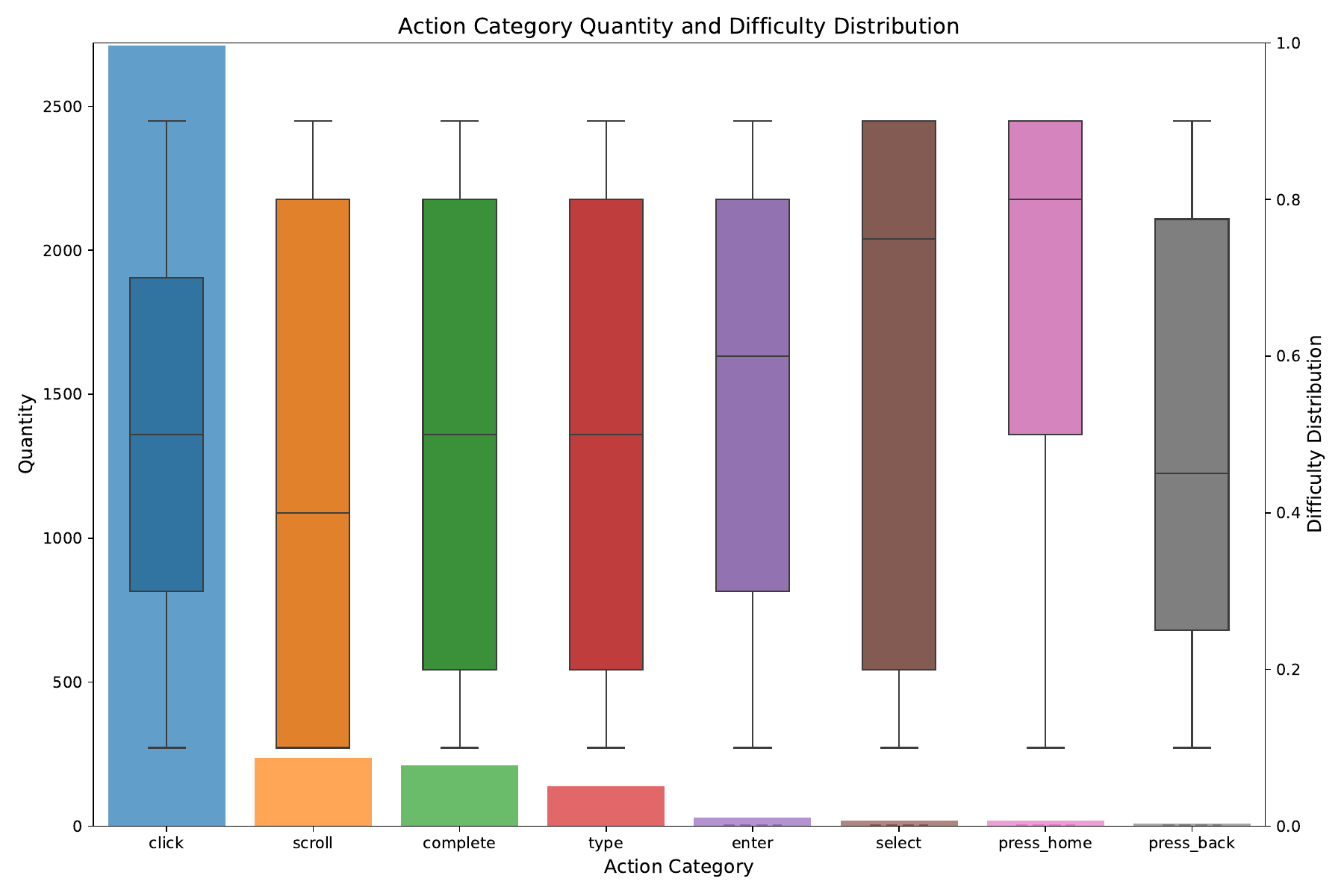}
    \caption{\scriptsize Action category quantity and difficulty distribution.}
    \label{fig:action_data_dist}
\end{subfigure}
\caption{Illustrations of image and action category quantity and difficulty distributions in the dataset GUI-R1-3K.}
\label{fig:data_dist}
\end{figure}

\noindent \textbf{Data filtering.} To filter out low-quality data for efficient RFT, we use the Qwen2.5VL-7B~\cite{qwen2.5vl} model to generate 10 responses for each example and evaluate them using a rule-based reward function designed for unified action space modeling. We remove the problems with an estimated accuracy of 0 or 1 to ensure a stable training process, resulting in 140K low-level data and 1.5K high-level data. Since the quantity of low-level data far exceeds that of high-level data, we randomly sample 1.5K low-level data and combine it with all high-level data to create a balanced dataset of 3K high-quality training examples, named GUI-R1-3K. The distribution of image categories, action types, and corresponding difficulty levels is demonstrated in Figure~\ref{fig:data_dist}.

\section{Experiments}

\subsection{Implementation Details}
\noindent \textbf{Training and inference details.}  For supervised fune-tuning~(SFT), we use the QwenVL2.5-3B/7B~\cite{qwen2.5vl} model as the base model for experiments and employ the LLaMA Factory~\cite{zheng2024llamafactory} framework for one epoch of training to avoid overfitting. For RFT, we use the EasyR1~\cite{zheng2025easyr1} framework for training over nine epochs. During inference, to ensure fairness, we apply a unified and simple prompt across all comparison methods, and conduct experiments under zero-shot prompt configurations. All experiments are conducted using 8$\times$NVIDIA A100-80G GPUs.

\noindent \textbf{Evaluation benchmarks.} We evaluate our model on eight agent benchmarks on three different platforms, including AndroidControl-Low~\cite{androidcontrol},  AndroidControl-High~\cite{androidcontrol}, GUI-Odyssey~\cite{guiodyssey}, ScreenSpot~\cite{seeclick}, ScreenSpot-Pro~\cite{li2024screenspot-pro}, GUI-Act-Web~\cite{guiact}, OmniAct-Web~\cite{omniact}, and OmniAct-Desktop~\cite{omniact}. We only use the test splits of these benchmarks for evaluation.

\noindent \textbf{Evaluation metrics.} Following Os-Atlas~\cite{osatlas}, we use three commonly adopted metrics for GUI agents in evaluation: action type prediction accuracy, click point prediction accuracy, and step success rate, denoted as Type, Grounding, and SR, respectively. In more detail, Type measures the exact match score between the predicted action types (e.g., `click' and `scroll') and the ground truth. Grounding evaluates the performance of GUI grounding in downstream tasks. Besides, SR represents the step-wise success rate, where a step is deemed successful only if both the predicted action and its associated arguments (e.g., point for click actions and input text for scroll actions) are correct.

\subsection{Experimental Results}
We here evaluate our \name model by comparing it with current state-of-the-art (SOTA) models on various tasks including GUI grounding tasks, GUI low-level tasks, and GUI high-level tasks.

\noindent \textbf{Grounding capability.} We evaluate the grounding capability of \name using  ScreenSpot~\cite{seeclick} and ScreenSpot-Pro~\cite{li2024screenspot-pro}. ScreenSpot assesses GUI grounding performance across mobile, desktop, and web platforms, while ScreenSpot-Pro focuses on high-resolution professional environments, featuring expert-annotated tasks spanning 23 applications, five industries, and three operating systems.

As shown in Table~\ref{tab:grounding}, compared to the previous SOTA model Os-Atlas-7B, which was trained with large-scale data using supervised fine-tuning (SFT), the RFT approach achieves superior performance on the 3B-sized Qwen2.5-VL model using only 0.2\% of the data (3K vs. 14M). Furthermore, compared to the base models QwenVL2.5-3B/7B and the SFT-trained QwenVL2.5* 3B/7B models using the same dataset, the RFT-based \name demonstrates significantly better performance in GUI grounding tasks. Moreover, at the 3B scale, \name achieves substantial gains over SFT models on ScreenSpot (80.08 vs. 63.55) and ScreenSpot-Pro (25.23 vs. 13.80), representing improvements of 26.3\% and 82.8\%, respectively. This highlights the effectiveness of the RL training framework in leveraging small-scale datasets to achieve significant performance improvements, which demonstrates its potential as a data-efficient and scalable approach for model training in resource-constrained environments.

\setlength{\tabcolsep}{1.5mm}{
\begin{table}[t] % 开始 table 环境
\centering
\caption{GUI grounding accuracy on ScreenSpot and ScreenSpot-Pro. All experiments are conducted under the same zero-shot prompt for fair comparison. * denotes supervised fine-tuned on GUI-R1-3K. The best results are in bold.} 
\resizebox{\linewidth}{!}{
\begin{tabular}{l|cc cc cc cc cc cc|cc cc}
\toprule
\multirow{3}{*}{Models} & \multicolumn{12}{c}{ScreenSpot-Pro} & \multicolumn{4}{c}{ScreenSpot}  \\
 & \multicolumn{2}{c}{Dev} & \multicolumn{2}{c}{Creative} & \multicolumn{2}{c}{CAD} & \multicolumn{2}{c}{Scientific} & \multicolumn{2}{c}{Office} & \multicolumn{2}{c}{OS} &  \multicolumn{2}{c}{Web} & \multicolumn{2}{c}{Desktop} \\
  & Text & Icon & Text & Icon & Text & Icon & Text & Icon & Text & Icon & Text & Icon & Text & Icon & Text & Icon \\
\midrule
\multicolumn{17}{l}{Supervised Fine-Tuning}\\
\midrule
SeeClick & 0.6 & 0.0 & 1.0 & 0.0 & 2.5 & 0.0 & 3.5 & 0.0 & 1.1 & 0.0 & 2.8 & 0.0 & 55.7 & 32.5 & 72.2 & 30.0 \\
Os-Atlas-4B & 7.1 & 0.0 & 3.0 & 1.4 & 2.0 & 0.0 & 9.0 & 5.5 & 5.1 & 3.8 & 5.6 & 0.0 & 82.6 & 63.1 & 72.1  & 45.7 \\
ShowUI-2B & 16.9 & 1.4 & 9.1 & 0.0 & 2.5 & 0.0 & 13.2 & 7.3 & 15.3 & 7.5 & 10.3 & 2.2 &  - &  - &  - &  - \\
CogAgent-18B & 14.9 & 0.7 & 9.6 & 0.0 & 7.1 & 3.1 & 22.2 & 1.8 & 13.0 & 0.0 & 5.6 & 0.0 &  70.4 &  28.6 &  74.2 & 20.0  \\
Aria-GUI & 16.2 & 0.0 & 23.7 & 2.1 & 7.6 & 1.6 & 27.1 & 6.4 & 20.3 & 1.9 & 4.7 & 0.0 &  - & -  &  - &  - \\
UGround-7B & 26.6 & 2.1 & 27.3 & 2.8 & 14.2 & 1.6 & 31.9 & 2.7 & 31.6 & 11.3 & 17.8 & 0.0 &   80.4 &  70.4 &  82.5 & 63.6  \\
Claude** & 22.0 & 3.9 & 25.9 & 3.4 & 14.5 & 3.7 & 33.9 & 15.8 & 30.1 & 16.3 & 11.0 & 4.5 & - & -  &  - &  - \\
Os-Atlas-7B & 33.1 & 1.4 & 28.8 & 2.8 & 12.2 & 4.7 & 37.5 & 7.3 & 33.9 & 5.7 & 27.1 & 4.5 &  90.8 &  74.2 &  91.7 & 62.8  \\
QwenVL2.5-3B* &  20.3 &  1.8 & 24.6  & 2.8 & 11.2 & 4.7 & 39.5  & 6.4  & 28.6 & 5.7  & 17.8  & 2.2 &  73.0  & 48.5  &  85.7 & 46.2  \\
QwenVL2.5-7B* &  31.4 & 1.8  &  27.3 & 3.5  & 15.7  & 5.1  &  40.7 & 7.9  &  39.7 & 8.9  & 32.4  &  6.9 &  87.8 & 68.2  &  90.3 &  62.8 \\
\midrule
\multicolumn{17}{l}{Zero Shot}\\
\midrule
QwenVL-7B & 0.0 & 0.0 & 0.0 & 0.0 & 0.0 & 0.0 & 0.7 & 0.0 & 0.0 & 0.0 & 0.0 & 0.0 & - & - & - & - \\
GPT-4o & 1.3 & 0.0 & 1.0 & 0.0 & 2.0 & 0.0 & 2.1 & 0.0 & 1.1 & 0.0 & 0.0 & 0.0 &  - & - & -  & -   \\
QwenVL2.5-3B & 16.2  & 1.4 &  23.3 &  1.4 &  10.2 & 4.7 &  38.2 & 6.4 & 24.3 & 3.8 & 15.0 & 1.1 &  60.8 & 43.5  & 70.1  &  35.0 \\
QwenVL2.5-7B &  33.1 & 2.1 & 23.7  & 3.5  & 12.2 & 6.3 & 36.8  &  7.3 & 37.8 & 7.5  & 30.8 & 6.9 & 86.9 & 65.1  & 89.7  & 60.0  \\
\midrule
\multicolumn{17}{l}{Reinforcement Fine-Tuning}\\
\midrule
UI-R1-3B &  22.7  & 4.1  &  27.3 &  3.5 &  11.2 &  6.3 & 43.4 & 11.8 & 32.2 & 11.3 & 13.1 & 4.5 &  85.2 & 73.3  & 90.2  & 59.3 \\
GUI-R1-3B & 33.8 & \textbf{4.8} &  \textbf{40.9}  & 5.6  &  \textbf{26.4} &  \textbf{7.8} &  \textbf{61.8} &  \textbf{17.3} & 53.6  &  17.0 & 28.1 & 5.6  & 89.6 & 72.1 &  \textbf{93.8} &  64.8 \\
GUI-R1-7B &  \textbf{49.4} & 4.8 & 38.9 & \textbf{8.4} &  23.9 & 6.3 &  55.6 & 11.8  &  \textbf{58.7} &  \textbf{26.4} & \textbf{42.1}  & \textbf{16.9} & \textbf{91.3}  & \textbf{75.7}  & 91.8  &  \textbf{73.6} \\
\bottomrule
\end{tabular}}
\label{tab:grounding}
\end{table} % 结束 table 环境
}
\noindent \textbf{Low-level task capability.} We evaluate the low-level task execution capability of \name using four benchmark datasets: AndroidControl-Low~\cite{androidcontrol}, GUI-Act-Web~\cite{li2024screenspot-pro}, OmniAct-Web, and OmniAct-Desktop~\cite{omniact}. AndroidControl-Low evaluates low-level task execution on mobile platforms, while GUI-Act-Web and OmniAct-Web focus on low-level task execution on web platforms. OmniAct-Desktop is used to test low-level task execution on desktop platforms.

As demonstrated in Table~\ref{tab:low_level}, our method impressively improves the success rate of GUI low-level tasks for 3B and 7B models, with the average success rate increasing from 55.65 to 80.88 at the 3B scale. Compared to UI-R1~\cite{ui-r1}, which is concurrent work also trained using RFT, our model achieves a 10-point improvement at the 3B scale, validating that RL training focused on high-level tasks can further enhance the model's understanding of low-level instructions. Note that an interesting observation is that the use of small-scale SFT data even leads to performance degradation on some metrics such as GR on AndroidControl-Low. This limitation stems from SFT's reliance on task-specific labeled data, which constrains the model's ability to adapt to unseen environments. In contrast, our RFT method not only enhances out-of-distribution~(OOD) generalization by optimizing task-specific rewards but also achieves this with fewer training examples, which provides a scalable and efficient alternative to traditional SFT methods.

\setlength{\tabcolsep}{1.5mm}{
\begin{table}[t] % 开始 table 环境
\centering
\caption{GUI low-level task accuracy on GUI-Act-Web, OmniAct-Web, OmniAct-Desktop, and AndroidControl-Low. All experiments are conducted under the same zero-shot prompt for fair comparison. * denotes supervised fine-tuned on GUI-R1-3K. The best results are in bold.} % caption 放在表格中
\resizebox{\linewidth}{!}{
\begin{tabular}{l|ccc ccc ccc ccc|c}
\toprule
\multirow{2}{*}{Models} & \multicolumn{3}{c}{GUI-Act-Web} & \multicolumn{3}{c}{OmniAct-Web} & \multicolumn{3}{c}{OmniAct-Desktop} & \multicolumn{3}{c}{AndroidControl-Low} & \multirow{2}{*}{Overall}  \\
 & Type & GR & SR & Type & GR & SR & Type & GR & SR & Type & GR & SR &  \\
\midrule
\multicolumn{14}{l}{Supervised Fine-Tuning}\\
\midrule

Os-Atlas-4B & 79.22 & 58.57 & 42.62 & 46.74 & 49.24 & 22.99 & 63.30 & 42.55 & 26.94 & 64.58 & 71.19 & 40.62 & 50.71 \\
Os-Atlas-7B & 86.95 & 75.61 & 57.02 & 85.63 & 69.35 & 59.15 & 90.24 & 62.87 & 56.73 & 73.00 & 73.37 & 50.94 & 70.07 \\

QwenVL2.5-3B* & 76.95 & 66.34 & 61.69 &  66.24 & 56.91 & 53.02 & 77.62 & 62.54 & 63.76 &  71.08 & 74.53 & 58.79 & 65.79 \\
QwenVL2.5-7B* & 87.66 & 84.77 & 79.89 & 81.62 & 73.45 & 73.39 &  86.23 & 80.17 & 79.80 & 84.00 & 85.74 & 64.32 & 80.09\\
\midrule
\multicolumn{14}{l}{Zero Shot}\\
\midrule
GPT-4o &  77.09 & 45.02 & 41.84 & 79.33 & 42.79 & 34.06 & 79.97 & 63.25 & 50.67 & 74.33 & 38.67 & 28.39 & 54.46 \\
QwenVL2.5-3B & 56.10 & 64.28 & 55.61 & 50.63 & 46.89 & 47.02 & 56.95 & 47.97 & 46.89 & 62.03 & 74.07 & 59.32 & 55.65\\
QwenVL2.5-7B & 86.59 & 84.39 & 78.63 & 79.15 &  71.32 & 71.21 &  84.74 & 79.89 & 79.66 & 83.44 & \textbf{87.08} & 62.50 & 79.05\\
\midrule
\multicolumn{14}{l}{Reinforcement Fine-Tuning}\\
\midrule
UI-R1-3B & 75.89 & 79.43 & 67.31 & 75.42 & 61.35 & 61.33 & 73.41 &  64.12 &  63.98 & 79.15 & 82.41 & 66.44 &  70.85 \\
GUI-R1-3B & 89.86 & 87.42 & 76.31 & 88.58 & 75.10 & 75.08 & 91.86  & 78.37 & 78.31 & 83.68 & 81.59 & 64.41 &  80.88 \\
GUI-R1-7B & \textbf{90.85} & \textbf{88.06} & \textbf{80.31} & \textbf{91.16} &  \textbf{77.29} & \textbf{77.35} & \textbf{92.20} & \textbf{83.36} & \textbf{83.33} & \textbf{85.17} & 84.02 & \textbf{66.52} & \textbf{83.30} \\
\bottomrule
\end{tabular}}
\label{tab:low_level}
\end{table} % 结束 table 环境
}

\noindent \textbf{High-level task capability.} We evaluate the high-level task execution capability of \name using AndroidControl-High~\cite{androidcontrol} and GUI-Odyssey~\cite{guiodyssey}. AndroidControl-High evaluates high-level task execution on mobile platforms, while GUI-Odyssey focuses on cross-app navigation scenarios, featuring high-level tasks spanning six applications and 203 apps.

As shown in Table~\ref{tab:high_level}, due to our unified action space with rule-based reward modeling, \name achieves SOTA on high-level GUI tasks. Compared to the closed-source model GPT-4o, our 3B-scale method achieves an absolute improvement of 21.06, demonstrating that RFT, in contrast to SFT, can efficiently and reliably enhance the success rate of GUI agents in real-world tasks. Furthermore, compared to UI-R1~\cite{ui-r1}, which focuses on improving low-level grounding capabilities, our model achieves an average improvement of 3.4 points at the 3B scale, with a particularly notable 27.2\% lead in the step success rate on GUI-Odyssey. This indicates that RL training focused on low-level tasks is insufficient for handling complex high-level instructions. RFT designed for high-level tasks is better suited as a direction for developing GUI agent models.

\setlength{\tabcolsep}{2.5mm}{
\begin{table}[t] % 开始 table 环境
\centering
\caption{GUI high-level task accuracy on AndroidControl-High and GUI-Odyssey. All experiments are conducted under the same zero-shot prompt for fair comparison. * denotes supervised fine-tuned on GUI-R1-3K. The best results are in bold.} % caption 放在表格中
\resizebox{0.85\linewidth}{!}{
\begin{tabular}{l|ccc ccc|c}
\toprule
\multirow{2}{*}{Models} &  \multicolumn{3}{c}{AndroidControl-High} & \multicolumn{3}{c}{GUI-Odyssey} & \multirow{2}{*}{Overall}  \\
 & Type & GR & SR & Type & GR & SR &  \\
\midrule
\multicolumn{8}{l}{Supervised Fine-Tuning}\\
\midrule

OS-Atlas-4B & 49.01 & 49.51 & 22.77 & 49.63 & 34.63 & 20.25 &  37.63 \\
OS-Atlas-7B &  57.44 & 54.90 & 29.83 & 60.42 & 39.74 & 26.96 &  44.88
\\
QwenVL2.5-3B* & 52.05 & 49.53 & 41.22 & 43.69 & 32.21 & 27.31 & 41.00 \\
QwenVL2.5-7B* & 69.15 & 58.69 & 48.11 & 56.78 & 38.65 & 34.44 & 50.97\\
\midrule
\multicolumn{8}{l}{Zero Shot}\\
\midrule
GPT-4o & 63.06 & 30.90 & 21.17 & 37.50 & 14.17 & 5.36 &   28.69\\
QwenVL2.5-3B & 47.81 & 46.51 & 38.90 & 37.40  & 26.49 & 26.69 &  37.30 \\
QwenVL2.5-7B & 68.67 & 59.71 & 47.06 & 55.60 & 37.78 & 34.37 &  50.53 \\
\midrule
\multicolumn{8}{l}{Reinforcement Fine-Tuning}\\
\midrule
UI-R1-3B & 57.85 & 55.70 & 45.44 &  52.16 & 34.46 & 32.49 &  46.35 \\
GUI-R1-3B & 58.04 & 56.24 & 46.55  & 54.84 & 41.52 & \textbf{41.33} &  49.75 \\
GUI-R1-7B & \textbf{71.63} & \textbf{65.56} & \textbf{51.67} &  \textbf{65.49} & \textbf{43.64} & 38.79 & \textbf{56.13} \\
\bottomrule
\end{tabular}}
\label{tab:high_level}
\end{table} % 结束 table 环境
}
\vspace{-6pt}
\subsection{Ablation Study}

\begin{figure}[t]
    \centering
    \includegraphics[width=0.64\textwidth]{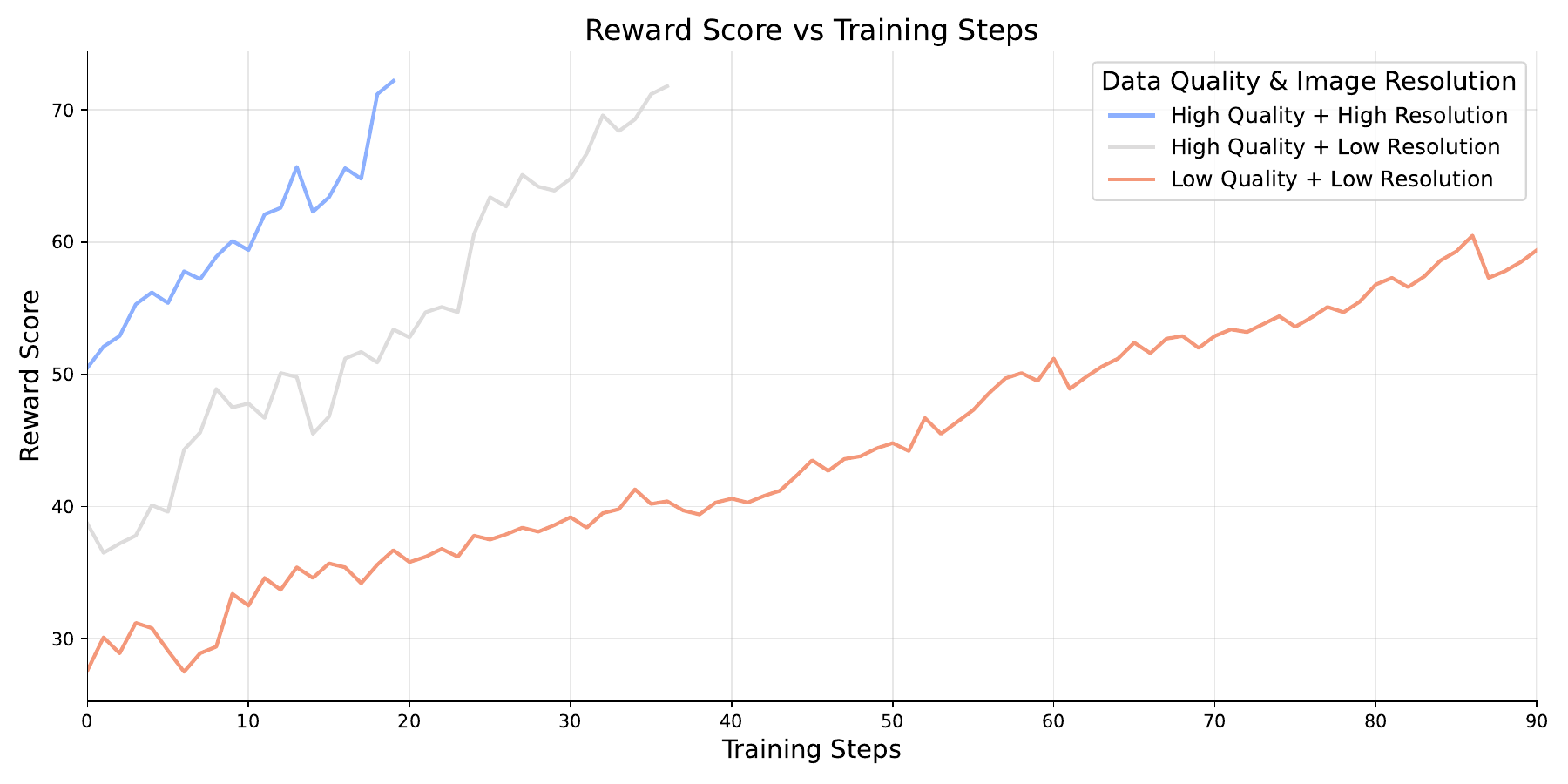}
    \vspace{-1em}
    \caption{Ablation study of image resolution and data quality.}
    \label{fig:hd_data_abaltion}
\end{figure}

\noindent \textbf{Image resolution and data quality.} To investigate the impact of image resolution and data quality on GUI RFT, we conduct corresponding ablation experiments, with the results shown in Figure~\ref{fig:hd_data_abaltion}. As observed, when using the filtered GUI-R1-3K dataset, the model requires only a few updates to achieve relatively high rewards. In contrast, training with unfiltered and low-quality data necessitates significantly more training time for the model to converge, with a noticeably lower performance ceiling. To further explore the effect of image resolution on model training, we increase the image resolution to twice its original size (from 1,048,576 pixels to 2,097,152 pixels). As shown in Figure~\ref{fig:hd_data_abaltion}, because of the high resolution of GUI task images and the small size of many UI elements, increasing the image resolution allows the model to perceive these elements more clearly, which accelerates the convergence speed of RFT and improves the performance ceiling.

\noindent \textbf{Coefficients in the reward function.} To explore the impact of the coefficients for format rewards and accuracy rewards in the reward function on the final performance, we conduct relevant ablation experiments, as shown in Table~\ref{tab:abla_coeffient}. The results indicate that reducing the coefficient ratio of format rewards leads to consistent performance improvements. This is because format rewards are easier to learn during training and often converge early in the process. By amplifying the accuracy rewards, the advantages of providing correct answers are further emphasized, ultimately leading to more performance improvements. 

\vspace{-6pt}

\setlength{\tabcolsep}{1.5mm}{
\begin{table}[h] % 开始 table 环境
\centering
\caption{Ablation study of the coeffient $\alpha$ and $\beta$ in reward function.} % caption 放在表格中
\resizebox{0.65\linewidth}{!}{
\begin{tabular}{ll|ccc ccc|c}
\toprule
\multirow{2}{*}{$\alpha$} &  \multirow{2}{*}{$\beta$} & \multicolumn{3}{c}{AndroidControl-High} & \multicolumn{3}{c}{GUI-Odyssey} & \multirow{2}{*}{Overall}  \\
& & Type & GR & SR & Type & GR & SR &  \\
\midrule

0.2 & 0.8 & \textbf{58.04} & \textbf{56.24} & \textbf{46.55}  & \textbf{54.84} & \textbf{41.52} & \textbf{41.33} &  \textbf{49.75} \\
0.5 & 0.5 & 57.93 & 55.91 & 46.62 & 52.77 & 37.44 &  35.66 &  47.72 \\
0.8 & 0.2 & 57.85 & 55.70 & 45.44 &  52.16 & 34.46 & 32.49 &  46.48 \\
\bottomrule
\end{tabular}}
\label{tab:abla_coeffient}
\end{table} % 结束 table 环境
}

\subsection{Visualization}
\begin{figure}[h]
  \centering
  \begin{subfigure}[b]{0.49\textwidth}
      \includegraphics[width=\textwidth]{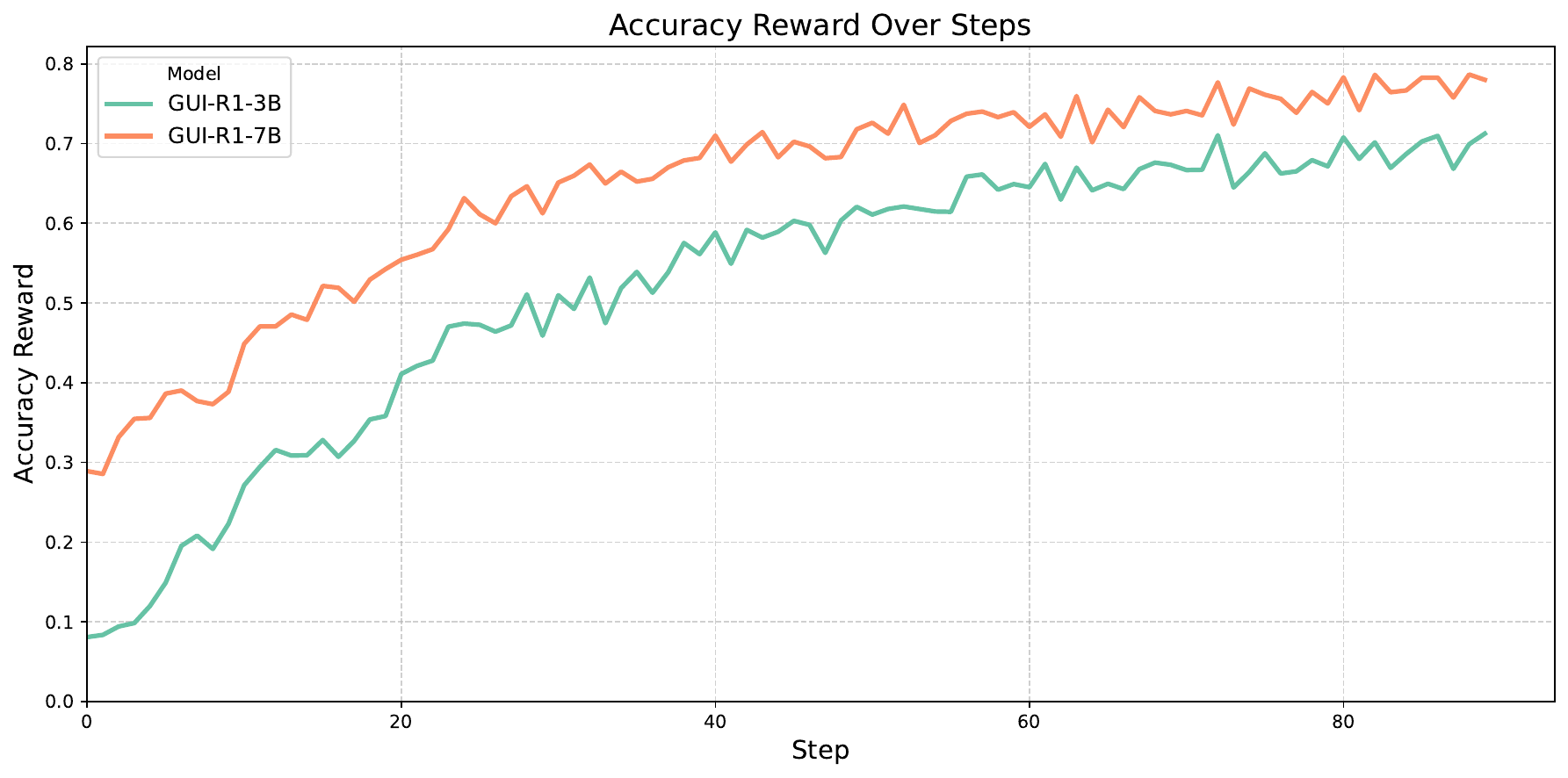}
      \caption{\scriptsize Accuracy reward curve with training steps.}
      \label{fig:acc_curve}
  \end{subfigure}
  \hfill
  \begin{subfigure}[b]{0.49\textwidth}
      \includegraphics[width=\textwidth]{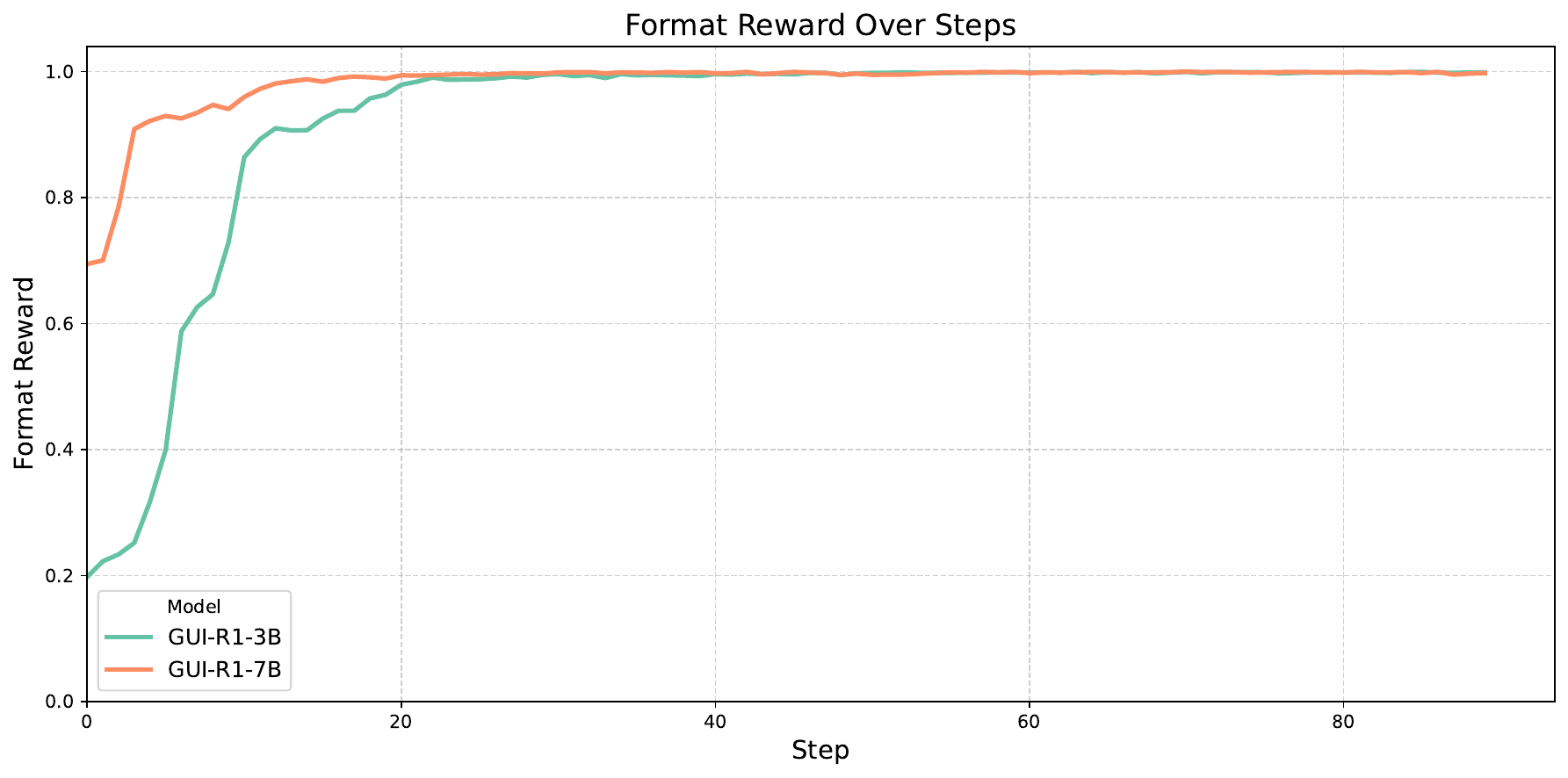}
      \caption{\scriptsize Format reward curve with training steps.}
      \label{fig:format_curve}
  \end{subfigure}
  \\
    \begin{subfigure}[b]{0.49\textwidth}
      \includegraphics[width=\textwidth]{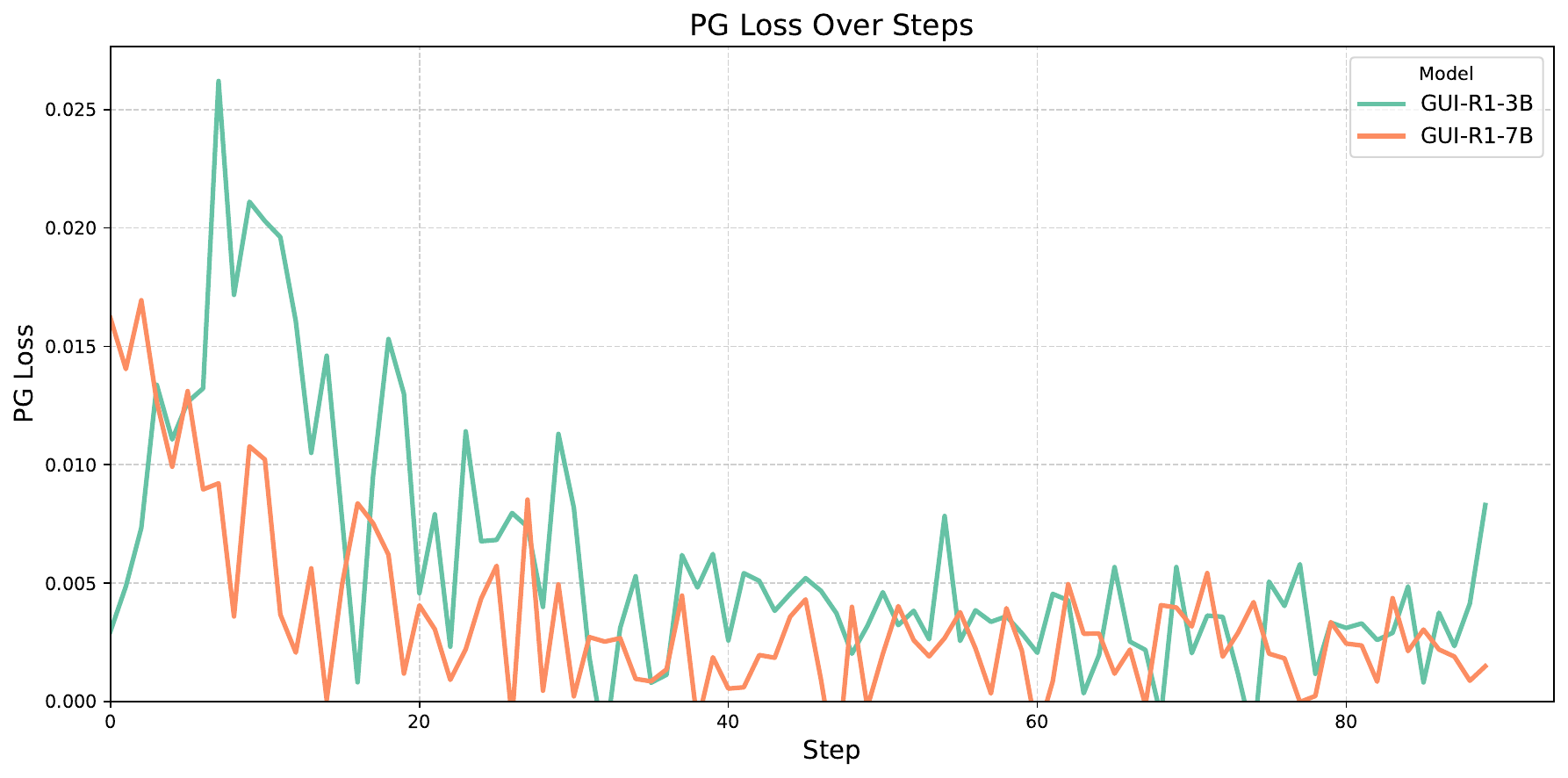}
      \caption{\scriptsize PG loss curve with training steps.}
      \label{fig:pgloss_curve}
  \end{subfigure}
  \hfill
  \begin{subfigure}[b]{0.49\textwidth}
      \includegraphics[width=\textwidth]{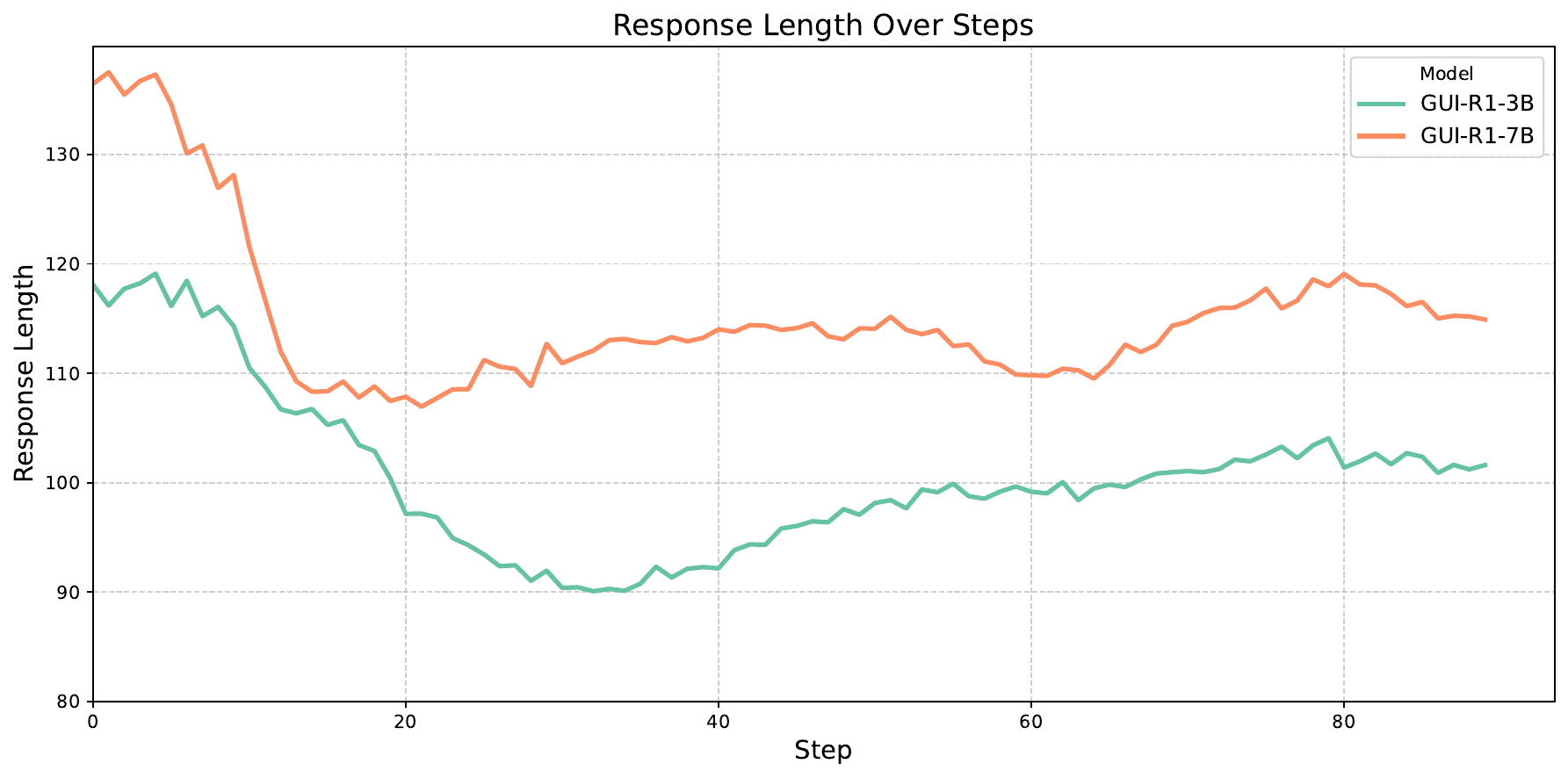}
      \caption{\scriptsize Mean response length curve with training steps.}
      \label{fig:rep_len_curve}
  \end{subfigure}
  \caption{\textbf{Visualization of the training process of \name.} To provide more details, we report the curves of \name's key metrics during training, including format reward, accuracy reward, mean response length, and policy gradient (PG) loss, as they vary with the training steps.}
  \label{fig:training_curve}
  \end{figure}

In Figure~\ref{fig:training_curve}, we provide additional visualization of the training process. As shown in Figure~\ref{fig:acc_curve} and Figure~\ref{fig:format_curve}, it can be observed that the format reward converges quickly in the early stages of training, while the accuracy reward becomes the main source of differentiated rewards in the later stages of training. Furthermore, as illustrated in Figure~\ref{fig:rep_len_curve}, the mean response length first decreases and then gradually increases, but the ``aha moment" does not occur. This may be due to the single-image input training method in a non-interactive environment, which prevents the model from autonomously tracing back the sequence of incorrect actions. Exploring multi-image high-level tasks in interactive environments could be a potential direction for inducing the emergence of the ``aha moment" in future research.

\section{Conclusion}
This paper presents GUI-R1, which is the first GUI reinforcement learning framework grounded in unified action space rule modeling. By integrating reinforcement fine-tuning with large vision-language models, GUI-R1 enables effective contextual action prediction and verifiable reward-driven learning in GUI environments. Extensive experiments demonstrate that GUI-R1 consistently outperforms baselines on various tasks. Moving forward, we plan to extend GUI-R1 to support collaborative multi-agent interaction and robust error correction policies, enabling the system to handle complex tasks with greater scalability.

\clearpage
\bibliography{reference}

\begin{thebibliography}{10}

\bibitem{osatlas}
Zhiyong Wu, Zhenyu Wu, Fangzhi Xu, Yian Wang, Qiushi Sun, Chengyou Jia, Kanzhi Cheng, Zichen Ding, Liheng Chen, Paul~Pu Liang, et~al.
\newblock Os-atlas: A foundation action model for generalist gui agents.
\newblock {\em arXiv preprint arXiv:2410.23218}, 2024.

\bibitem{uitars}
Yujia Qin, Yining Ye, Junjie Fang, Haoming Wang, Shihao Liang, Shizuo Tian, Junda Zhang, Jiahao Li, Yunxin Li, Shijue Huang, et~al.
\newblock Ui-tars: Pioneering automated gui interaction with native agents.
\newblock {\em arXiv preprint arXiv:2501.12326}, 2025.

\bibitem{seeclick}
Kanzhi Cheng, Qiushi Sun, Yougang Chu, Fangzhi Xu, Yantao Li, Jianbing Zhang, and Zhiyong Wu.
\newblock Seeclick: Harnessing gui grounding for advanced visual gui agents.
\newblock {\em arXiv preprint arXiv:2401.10935}, 2024.

\bibitem{qwen2.5vl}
Shuai Bai, Keqin Chen, Xuejing Liu, Jialin Wang, Wenbin Ge, Sibo Song, Kai Dang, Peng Wang, Shijie Wang, Jun Tang, et~al.
\newblock Qwen2. 5-vl technical report.
\newblock {\em arXiv preprint arXiv:2502.13923}, 2025.

\bibitem{visual-rft}
Ziyu Liu, Zeyi Sun, Yuhang Zang, Xiaoyi Dong, Yuhang Cao, Haodong Duan, Dahua Lin, and Jiaqi Wang.
\newblock Visual-rft: Visual reinforcement fine-tuning.
\newblock {\em arXiv preprint arXiv:2503.01785}, 2025.

\bibitem{visionr1}
Wenxuan Huang, Bohan Jia, Zijie Zhai, Shaosheng Cao, Zheyu Ye, Fei Zhao, Yao Hu, and Shaohui Lin.
\newblock Vision-r1: Incentivizing reasoning capability in multimodal large language models.
\newblock {\em arXiv preprint arXiv:2503.06749}, 2025.

\bibitem{r1v}
Liang Chen, Lei Li, Haozhe Zhao, Yifan Song, and Vinci.
\newblock R1-v: Reinforcing super generalization ability in vision-language models with less than \$3.
\newblock \url{https://github.com/Deep-Agent/R1-V}, 2025.
\newblock Accessed: 2025-02-02.

\bibitem{vlmr1}
Haozhan Shen, Zilun Zhang, Kangjia Zhao, Qianqian Zhang, Ruochen Xu, and Tiancheng Zhao.
\newblock Vlm-r1: A stable and generalizable r1-style large vision-language model.
\newblock \url{https://github.com/om-ai-lab/VLM-R1}, 2025.
\newblock Accessed: 2025-02-15.

\bibitem{ui-r1}
Zhengxi Lu, Yuxiang Chai, Yaxuan Guo, Xi~Yin, Liang Liu, Hao Wang, Guanjing Xiong, and Hongsheng Li.
\newblock Ui-r1: Enhancing action prediction of gui agents by reinforcement learning.
\newblock {\em arXiv preprint arXiv:2503.21620}, 2025.

\bibitem{deepseekr1}
Daya Guo, Dejian Yang, Haowei Zhang, Junxiao Song, Ruoyu Zhang, Runxin Xu, Qihao Zhu, Shirong Ma, Peiyi Wang, Xiao Bi, et~al.
\newblock Deepseek-r1: Incentivizing reasoning capability in llms via reinforcement learning.
\newblock {\em arXiv preprint arXiv:2501.12948}, 2025.

\bibitem{sumers2023cognitive}
Theodore Sumers, Shunyu Yao, Karthik Narasimhan, and Thomas Griffiths.
\newblock Cognitive architectures for language agents.
\newblock {\em Transactions on Machine Learning Research}, 2023.

\bibitem{wang2024survey}
Lei Wang, Chen Ma, Xueyang Feng, Zeyu Zhang, Hao Yang, Jingsen Zhang, Zhiyuan Chen, Jiakai Tang, Xu~Chen, Yankai Lin, et~al.
\newblock A survey on large language model based autonomous agents.
\newblock {\em Frontiers of Computer Science}, 18(6):186345, 2024.

\bibitem{sun2023corex}
Qiushi Sun, Zhangyue Yin, Xiang Li, Zhiyong Wu, Xipeng Qiu, and Lingpeng Kong.
\newblock Corex: Pushing the boundaries of complex reasoning through multi-model collaboration.
\newblock {\em arXiv preprint arXiv:2310.00280}, 2023.

\bibitem{uground}
Boyu Gou, Ruohan Wang, Boyuan Zheng, Yanan Xie, Cheng Chang, Yiheng Shu, Huan Sun, and Yu~Su.
\newblock Navigating the digital world as humans do: Universal visual grounding for gui agents.
\newblock {\em arXiv preprint arXiv:2410.05243}, 2024.

\bibitem{jaech2024openai}
Aaron Jaech, Adam Kalai, Adam Lerer, Adam Richardson, Ahmed El-Kishky, Aiden Low, Alec Helyar, Aleksander Madry, Alex Beutel, Alex Carney, et~al.
\newblock Openai o1 system card.
\newblock {\em arXiv preprint arXiv:2412.16720}, 2024.

\bibitem{shao2024deepseekmath}
Zhihong Shao, Peiyi Wang, Qihao Zhu, Runxin Xu, Junxiao Song, Xiao Bi, Haowei Zhang, Mingchuan Zhang, YK~Li, Y~Wu, et~al.
\newblock Deepseekmath: Pushing the limits of mathematical reasoning in open language models.
\newblock {\em arXiv preprint arXiv:2402.03300}, 2024.

\bibitem{code-r1}
Jiawei Liu and Lingming Zhang.
\newblock Code-r1: Reproducing r1 for code with reliable rewards.
\newblock {\em arXiv preprint arXiv:2503.18470}, 2025.

\bibitem{RAGEN}
Zihan Wang*, Kangrui Wang*, Qineng Wang*, Pingyue Zhang*, Linjie Li*, Zhengyuan Yang, Kefan Yu, Minh~Nhat Nguyen, Monica Lam, Yiping Lu, Kyunghyun Cho, Jiajun Wu, Li~Fei-Fei, Lijuan Wang, Yejin Choi, and Manling Li.
\newblock Training agents by reinforcing reasoning, 2025.

\bibitem{pan2025metaspatial}
Zhenyu Pan and Han Liu.
\newblock Metaspatial: Reinforcing 3d spatial reasoning in vlms for the metaverse.
\newblock {\em arXiv preprint arXiv:2503.18470}, 2025.

\bibitem{meng2025mm}
Fanqing Meng, Lingxiao Du, Zongkai Liu, Zhixiang Zhou, Quanfeng Lu, Daocheng Fu, Botian Shi, Wenhai Wang, Junjun He, Kaipeng Zhang, et~al.
\newblock Mm-eureka: Exploring visual aha moment with rule-based large-scale reinforcement learning.
\newblock {\em arXiv preprint arXiv:2503.07365}, 2025.

\bibitem{fineweb}
Guilherme Penedo, Hynek Kydl{\'\i}{\v{c}}ek, Anton Lozhkov, Margaret Mitchell, Colin~A Raffel, Leandro Von~Werra, Thomas Wolf, et~al.
\newblock The fineweb datasets: Decanting the web for the finest text data at scale.
\newblock In {\em NeurIPS}, pages 30811--30849, 2024.

\bibitem{uibert}
Chongyang Bai, Xiaoxue Zang, Ying Xu, Srinivas Sunkara, Abhinav Rastogi, Jindong Chen, and Blaise~Aguera y~Arcas.
\newblock Uibert: Learning generic multimodal representations for ui understanding, 2021.

\bibitem{amex}
Yuxiang Chai, Siyuan Huang, Yazhe Niu, Han Xiao, Liang Liu, Dingyu Zhang, Peng Gao, Shuai Ren, and Hongsheng Li.
\newblock Amex: Android multi-annotation expo dataset for mobile gui agents.
\newblock {\em arXiv preprint arXiv:2407.17490}, 2024.

\bibitem{li2020mapping}
Yang Li, Jiacong He, Xin Zhou, Yuan Zhang, and Jason Baldridge.
\newblock Mapping natural language instructions to mobile ui action sequences.
\newblock {\em arXiv preprint arXiv:2005.03776}, 2020.

\bibitem{zheng2024llamafactory}
Yaowei Zheng, Richong Zhang, Junhao Zhang, Yanhan Ye, Zheyan Luo, Zhangchi Feng, and Yongqiang Ma.
\newblock Llamafactory: Unified efficient fine-tuning of 100+ language models.
\newblock In {\em ACL}, 2024.

\bibitem{zheng2025easyr1}
Yaowei Zheng, Junting Lu, Shenzhi Wang, and Y~Xiong.
\newblock Easyr1: An efficient, scalable, multi-modality rl training framework, 2025.

\bibitem{androidcontrol}
Wei Li, William Bishop, Alice Li, Chris Rawles, Folawiyo Campbell-Ajala, Divya Tyamagundlu, and Oriana Riva.
\newblock On the effects of data scale on computer control agents.
\newblock {\em arXiv preprint arXiv:2406.03679}, 2024.

\bibitem{guiodyssey}
Quanfeng Lu, Wenqi Shao, Zitao Liu, Fanqing Meng, Boxuan Li, Botong Chen, Siyuan Huang, Kaipeng Zhang, Yu~Qiao, and Ping Luo.
\newblock Gui odyssey: A comprehensive dataset for cross-app gui navigation on mobile devices.
\newblock {\em arXiv preprint arXiv:2406.08451}, 2024.

\bibitem{li2024screenspot-pro}
Kaixin Li, Ziyang Meng, Hongzhan Lin, Ziyang Luo, Yuchen Tian, Jing Ma, Zhiyong Huang, and Tat-Seng Chua.
\newblock Screenspot-pro: Gui grounding for professional high-resolution computer use.
\newblock {\em Workshop on Reasoning and Planning for Large Language Models}, 2025.

\bibitem{guiact}
Wentong Chen, Junbo Cui, Jinyi Hu, Yujia Qin, Junjie Fang, Yue Zhao, Chongyi Wang, Jun Liu, Guirong Chen, Yupeng Huo, et~al.
\newblock Guicourse: From general vision language models to versatile gui agents.
\newblock {\em arXiv preprint arXiv:2406.11317}, 2024.

\bibitem{omniact}
Raghav Kapoor, Yash~Parag Butala, Melisa Russak, Jing~Yu Koh, Kiran Kamble, Waseem AlShikh, and Ruslan Salakhutdinov.
\newblock Omniact: A dataset and benchmark for enabling multimodal generalist autonomous agents for desktop and web.
\newblock In {\em ECCV}, pages 161--178. Springer, 2024.

\end{thebibliography}
\bibliographystyle{unsrt}

%%%%%%%%%%%%%%%%%%%%%%%%%%%%%%%%%%%%%%%%%%%%%%%%%%%%%%%%%%%%

\appendix

%%%%%%%%%%%%%%%%%%%%%%%%%%%%%%%%%%%%%%%%%%%%%%%%%%%%%%%%%%%%

\newpage

\end{document}